\documentclass[letterpaper, 10 pt, conference]{ieeeconf}  

\IEEEoverridecommandlockouts                              

\usepackage{amssymb}  

\usepackage{amsmath}
\usepackage{graphicx}
\usepackage{cite}

\title{\LARGE \bf
Acceleration-Based Control of Fixed-Wing UAVs for Guidance Applications
}

\author{Jixiang Wang$^{1}$, Siyuan Yang$^{1}$, Ziyi Wu$^{1}$, Siqi Wei$^{1}$, Ashay Wakode$^{2}$, Agata Barci\'s$^{2}$, Hung Nguyen$^{2}$ and Shaoming He$^{1}$ 
\thanks{*This work was supported by Technology Innovation Institute under Contract No. TII/ARRC/2154/2023.}
\thanks{$^{1}$Jixiang Wang, Siyuan Yang, Ziyi Wu, Siqi Wei and Shaoming He are with the School of Aerospace Engineering, Beijing Institute of Technology, Beijing 100081, China.}
\thanks{$^{2}$Ashay Wakode, Agata Barci\'s, and Hung Nguyen are with the Autonomous Robotics Research Centre, Technology Innovation Institute, P.O.Box: 9639, Masdar City, Abu Dhabi, United Arab Emirates.}
}

\begin{document}

\maketitle
\thispagestyle{empty}
\pagestyle{empty}

\begin{abstract}
Acceleration-commanded guidance laws (e.g., proportional navigation) are attractive for high-level decision making, but their direct deployment on fixed-wing UAVs is challenging because accelerations are not directly actuated and must be realized through attitude and thrust under flight-envelope constraints. This paper presents an acceleration-level outer-loop control framework that converts commanded tangential and normal accelerations into executable body-rate and normalized thrust commands compatible with mainstream autopilots (e.g., PX4/APM). For the normal channel, we derive an engineering mapping from the desired normal acceleration to roll- and pitch-rate commands that regulate the direction and magnitude of the lift vector under small-angle assumptions. For the tangential channel, we introduce an energy-based formulation inspired by total energy control and identify an empirical thrust–energy-acceleration relationship directly from flight data, avoiding explicit propulsion modeling or thrust bench calibration. We further discuss priority handling between normal and tangential accelerations under saturation and non-level maneuvers. Extensive real-flight experiments on a VTOL fixed-wing platform demonstrate accurate acceleration tracking and enable practical implementation of proportional navigation using only body-rate and normalized thrust interfaces.
\end{abstract}

\section{Introduction}

Fixed-wing unmanned aerial vehicles (UAVs) are increasingly deployed in long-range and long-endurance missions, including surveillance\cite{beard2006decentralized}, environmental monitoring\cite{rufino2024mission}, and logistics\cite{zhang2024design}, thanks to their high cruise efficiency and payload capacity. Despite these advantages, guidance and control for fixed-wing platforms remain challenging in practice due to underactuation, strong nonlinear and coupled dynamics, and sensitivity to disturbances (e.g., wind). These characteristics complicate the reliable execution of agile maneuvers, especially when guidance objectives are naturally expressed at the acceleration level.

A large class of guidance and autonomy algorithms—ranging from classical proportional navigation (PN) \cite{he2025review,adler1956missile,zarchan2012tactical,he2018optimality} to modern interception, collision \cite{lindqvist2020nonlinear}, and motion planning strategies—produce desired accelerations as their primary outputs. This acceleration-centric interface is attractive because it provides a compact and physically meaningful command that can be composed with higher-level decision-making. On highly maneuverable platforms (e.g., multirotor UAVs), acceleration commands can often be tracked directly or with relatively simple mappings to attitude and thrust \cite{Lee2019Robust}, enabling straightforward integration of acceleration-based guidance. In contrast, directly transferring this paradigm to fixed-wing UAVs is nontrivial: a fixed-wing aircraft must maintain sufficient airspeed to generate lift, and its achievable acceleration is constrained by aerodynamic characteristics and flight-envelope limits. As a result, acceleration cannot be commanded independently in arbitrary directions; instead, it must be realized indirectly through coordinated changes in attitude and propulsion.

This mismatch reveals a key engineering bottleneck: how to transform acceleration commands generated by a guidance layer into executable commands compatible with practical fixed-wing flight-control stacks. Most widely used autopilots provide high-rate tracking for low-level commands such as body angular rates and a normalized thrust input, but they do not natively expose an acceleration-tracking interface. Consequently, an acceleration-level outer-loop that respects fixed-wing physics while remaining compatible with existing inner-loop controllers is needed to bridge high-level acceleration commands and low-level actuation.

This paper proposes an acceleration-command realization framework for fixed-wing UAVs that enables acceleration-based guidance laws to be implemented without modifying the inner-loop flight controller. The key idea is to decompose the desired acceleration into tangential and normal components and generate corresponding attitude angular-rate and normalized thrust commands that can be tracked by standard autopilots. For the normal component, the method constructs a physically interpretable mapping from the desired normal acceleration to roll- and pitch-rate commands, leveraging the relationship between lift-vector orientation/magnitude and lateral/vertical acceleration. For the tangential component, we adopt an energy-based perspective inspired by total-energy control concepts \cite{lambregts1983vertical} and develop a practical in-flight identification procedure to relate normalized thrust to an equivalent energy acceleration quantity—thereby avoiding reliance on an accurate propulsion model or thrust bench calibration. Finally, because fixed-wing platforms inevitably encounter saturations and coupling between normal and tangential dynamics (e.g., during climbs/descents or aggressive maneuvers), we incorporate a simple and implementable priority handling mechanism to trade off normal-acceleration tracking against tangential-acceleration tracking based on mission needs. The main contributions of this work are summarized as follows: 
\begin{enumerate}
    \item \textbf{Acceleration-to-command mapping for fixed-wing UAVs}: an acceleration-level outer-loop that converts desired normal and tangential accelerations into executable body angular-rate and normalized thrust commands compatible with mainstream fixed-wing autopilots.
    \item \textbf{Model-light thrust realization via in-flight identification}: an energy-based method to establish the thrust–acceleration relationship directly from flight data, avoiding explicit propulsion modeling while remaining suitable for real-time use.
    \item \textbf{Engineering validation on a real platform}: real-world flight experiments demonstrating acceleration tracking performance and enabling the practical implementation of a representative acceleration-based guidance law (proportional navigation) on a fixed-wing UAV.
\end{enumerate}

\section{Problem Description}\label{sec:problem}

This section formulates the acceleration-level outer-loop design problem for a fixed-wing UAV under practical implementation constraints. We first define the coordinate frames and modeling assumptions used throughout the paper, and then state the acceleration-tracking objective together with the actuation interfaces available in typical fixed-wing autopilots.

\begin{table}[!t] 
\renewcommand{\arraystretch}{1.3}
\caption{Nomenclature}
\label{table_nomenclature}
\centering

\begin{tabular}{@{}p{0.16\columnwidth} p{0.75\columnwidth}@{}}
	$F^i$ 
	& Inertial reference frame using the North--East--Down (NED) convention \\
	
	$F^{v2}$ 
	& Intermediate reference frame obtained by rotating $F^i$ through yaw and pitch angles \\
	
	$F^b$ 
	& Body-fixed frame attached to the center of mass of the vehicle \\
	
	$F^v$ 
	& Velocity frame \\
	
	$x^*, y^*, z^*$ 
	& Axes of the reference frame $F^*$ \\

	$\mathbf{e}_x^*, \mathbf{e}_y^*, \mathbf{e}_z^*$ 
	& Unit vectors along the positive $x$-, $y$-, and $z$-axes of frame $F^*$, respectively \\
	
	$x_x^*, x_y^*, x_z^*, $
	& The component of vector $\mathbf{x}$ along the $x$-, $y$-, and $z$-axes of the coordinate frame $F^*$, respectively; $x_x^*=\mathbf{x}\cdot \mathbf{e}_x^*$\\
	
	$R_x, R_y, R_z$ 
	& Elementary rotation matrices about the $x$-, $y$-, and $z$-axes, respectively \\
	
	$R^{*}_{**}$ 
	& Rotation matrix from frame $F^{**}$ to frame $F^*$ \\
	
	$\phi, \theta, \psi$ 
	& Euler angles (roll, pitch, yaw) \\
	
	$p, q, r$ 
	& Roll, pitch and yaw rate \\
	
	$\alpha, \beta$ 
	& Angle of attack and sideslip angle \\
	
	$\mathbf{a}$
	& Acceleration vector\\ 
	
	$\mathbf{g}$ 
	& Gravitational acceleration; $g = \| \mathbf{g} \|$ \\
	
	$\mathbf{V}$ 
	& Ground velocity vector; $V = \| \mathbf{V} \|$ \\
	
	$V_a$ 
	& Airspeed \\
	
	$T$ 
	& Thrust force \\
	
	$D$
	& Aerodynamic drag \\
	
	$\rho$
	& Air density \\
	
	$S$
	& Reference Area \\
	
	$C_D$
	& Drag Coefficient \\
	
	$N$
	& Proportional navigation gain \\
	
	$V_{cl}$
	& Closing speed between the UAV and the target \\
	
	$\lambda$
	& Line-of-sight \\
	
	$(\cdot)_c$ 
	& Commanded value of a variable
\end{tabular}
\end{table}

\subsection{Coordinate Frames and Modeling Assumptions}

The UAV attitude is parameterized by Euler angles $(\phi,\theta,\psi)$ (roll, pitch, yaw). The inertial frame $F^i$ follows the North--East--Down (NED) convention, and the intermediate frame $F^{v2}$, velocity frame $F^v$, and body frame $F^b$ are defined in Table~\ref{table_nomenclature}~\cite{beard2012small}. The rotations between different frames are written as
\begin{equation}
\begin{aligned}
	R_i^{v2} &= R_y(\theta) R_z(\psi),  \\
	R_{v2}^b &= R_x(\phi),  \\
	R_b^v &= R_y(-\alpha) R_z(\beta). 
\end{aligned}
\end{equation}

To keep the analysis focused on implementable acceleration-command realization, the following assumptions are adopted: (i) the UAV operates with small angle of attack and sideslip during cruise and guidance maneuvers, and does not execute large-attitude maneuvers (i.e., $|\phi|$ and $|\theta|$ remain below $30^\circ$); (ii) the flight speed is well below the speed of sound, so compressibility effects are neglected; and (iii) environmental wind is small relative to the UAV airspeed and is neglected in this study.

In practical implementations, the velocity direction and the corresponding velocity frame $F^v$ are difficult to measure directly. The angles of attack and sideslip are typically estimated and are subject to noise and modeling uncertainty. Since this paper targets a realizable implementation in existing fixed-wing flight-control stacks, the intermediate frame $F^{v2}$ and the body frame $F^b$ are used for acceleration decomposition and control design. Under the small-$\alpha$ and small-$\beta$ assumptions, $F^b$ and $F^v$ are approximately aligned, which makes this approximation suitable for practical guidance realizations.

\subsection{Acceleration-Level Control Problem}

Assume that a high-level guidance module provides a desired acceleration command $\mathbf{a}_c \in \mathbb{R}^3$, which serves as the outer-loop reference. For a fixed-wing UAV, this command cannot be actuated directly; it must be generated through thrust modulation and attitude changes that redirect the lift vector, subject to aerodynamic characteristics and flight-envelope constraints.

Mainstream fixed-wing autopilots such as PX4~\cite{meier2015px4, saengphet2017implementation, Baldi2022ArduPilot-Based} and APM~\cite{bin2009design} typically provide high-bandwidth inner-loop tracking of body angular-rate commands and a normalized thrust input.  In this work, these inner loops are treated as ideal, and the low-level attitude  and thrust controllers are not redesigned~\cite{poksawat2016automatic} The core problem addressed here is to construct an acceleration-level outer loop that maps $\mathbf{a}_c$ to $(p_c,q_c,T_c)$ without requiring modifications to existing controller architectures.

A key practical difficulty is that the thrust command $T_c$ is dimensionless (typically in $[0,1]$), while the relationship between $T_c$ and the produced tangential force/acceleration depends on propulsion characteristics, flight condition, and the aerodynamic environment. Obtaining an accurate and unified analytical model is difficult, and direct thrust measurement is often unavailable. Therefore, the acceleration-command realization layer should be model-light and identifiable from flight data. The next section develops such a mapping based on acceleration decomposition and a total-energy formulation.

\section{Acceleration Tracking Control}\label{sec:accel_control}
Building on the problem statement in Section~\ref{sec:problem}, this section presents an acceleration-tracking outer loop tailored to practical fixed-wing flight-control stacks. The objective is to map the desired acceleration command provided by the guidance module into executable body angular-rate commands and a normalized thrust command, i.e., $(p_c,q_c,T_c)$.

Let the (ground) velocity vector be $\mathbf{V}=V\,\mathbf{e}_x^v$, where $\mathbf{e}_x^v$ is the unit vector along the $x$-axis of the velocity frame $F^v$. Differentiating yields
\begin{equation}
	\dot{\mathbf{V}} = \dot{V}\,\mathbf{e}_x^v + V\,\dot{\mathbf{e}}_x^v.
\end{equation}

Accordingly, the commanded acceleration $\mathbf{a}_c$ can be decomposed into a tangential component along the velocity direction, $\mathbf{a}_c^t := \dot{V}\,\mathbf{e}_x^v$, and a normal component orthogonal to the velocity direction, $\mathbf{a}_c^n := V\,\dot{\mathbf{e}}_x^v$. Under the small-$\alpha$ and small-$\beta$ assumptions, the body $x$-axis is approximately aligned with $\mathbf{e}_x^v$, which enables an implementable mapping from $(\mathbf{a}_c^n,\mathbf{a}_c^t)$ to $(p_c,q_c,T_c)$.

\subsection{Normal Acceleration Tracking}

This subsection establishes a mapping from the desired normal acceleration $\mathbf{a}_c^n$ to the roll-rate and pitch-rate commands $(p_c,q_c)$. For a fixed-wing UAV, normal acceleration is primarily generated by redirecting and scaling the lift vector. Neglecting drag and other secondary forces in the normal plane, the required specific force produced by lift is
\begin{equation}
	\mathbf{a}_L = \mathbf{a}_c^n + \mathbf{g}.
\end{equation}

In bank-to-turn (BTT) coordinated flight, the lift vector is rotated by the roll angle. Hence, the commanded roll angle can be chosen as the angle between the required lift vector and the negative vertical axis $\mathbf{e}_z^{v2}$ in the intermediate frame $F^{v2}$~\cite{yoshitani2010flight}:
\begin{equation}
	\phi_c = \angle(\mathbf{a}_L,\, -\mathbf{e}_z^{v2}).
\end{equation}

To achieve fast tracking of $\phi_c$, a proportional law is adopted:
\begin{equation}
	\dot{\phi}_c = k_\phi(\phi_c - \phi).
\end{equation}
where $k_\phi$ is the tuning gain. From rigid-body kinematics,
\begin{equation}
	\dot{\phi} = p + q\sin\phi\tan\theta + r\cos\phi\tan\theta.
\end{equation}

For moderate attitudes ($\theta\approx 0$), the coupling terms can be neglected and the roll rate is dominated by $p$, leading to the roll-rate command
\begin{equation}
	p_c = k_\phi(\phi_c - \phi).
\end{equation}

Next, we relate the pitch rate to the commanded normal acceleration magnitude. Using the vector derivative relation
\begin{equation}
	\dot{\mathbf{e}}_x^v = \boldsymbol{\omega}_{v/i}^v \times \mathbf{e}_x^v,
\end{equation}
where $\boldsymbol{\omega}_{v/i}^v = [p^v, q^v, r^v]^T$ is the angular velocity of $F^v$ with respect to $F^i$ expressed in $F^v$, we obtain
\begin{equation}
	V\,\dot{\mathbf{e}}_x^v
	= V
	\begin{bmatrix} p^v & q^v & r^v \end{bmatrix}^T
	\times
	\begin{bmatrix} 1 \\ 0 \\ 0 \end{bmatrix}
	=
	\begin{bmatrix} 0 \\ Vr^v \\ -Vq^v \end{bmatrix}.
\end{equation}

Therefore, the $z$-axis component of the normal acceleration in the velocity frame is directly related to $q^v$. Ignoring the small frame differences under the small-$\alpha$ and small-$\beta$ assumptions (so that $q^v \approx q$), the pitch-rate command is approximated as
\begin{equation}
	q_c = -\frac{(a_c^n)_z^v}{V},
\end{equation}
where $(a_c^n)_z^v$ denotes the $z$-axis component of $\mathbf{a}_c^n$ expressed in $F^v$.

The normal-acceleration mapping is derived under the small angle of attack and sideslip assumptions. 
If these assumptions are violated, minor steady-state tracking errors may appear due to unmodeled aerodynamic effects. 
In practical implementations, such errors can be reduced by introducing a simple integral term in the normal acceleration loop without altering the overall control structure.

With $p_c$ shaping the lift-vector direction and $q_c$ regulating the rate of change of the velocity direction, the proposed mapping provides a practical and physically interpretable mechanism to realize the desired normal acceleration $\mathbf{a}_c^n$ using only body-rate commands.

\subsection{Tangential Acceleration Tracking}

This subsection develops a mapping from the desired tangential acceleration to the normalized thrust command $T_c$ without requiring an explicit propulsion model. In practical fixed-wing flight-control systems (e.g., PX4 and APM), $T_c$ is a dimensionless normalized input in $[0,1]$, and its relationship to the produced thrust and tangential acceleration depends on propulsion characteristics, propeller efficiency, and flight condition. We therefore adopt a total-energy perspective and identify an empirical thrust--energy-acceleration relationship directly from flight data.

The total mechanical energy of the aircraft is
$E = mgh + \frac{1}{2}mV^2$.
where $m$ and $h$ are vehicle's mass and altitude, respectively. Taking the time derivative yields
\begin{equation}\label{dTEDefine}
	\dot{E} = -mgV_z^i + mV\dot{V},
\end{equation}
where $V_z^i$ is the downward velocity component in the NED inertial frame (so that $\dot{h}=-V_z^i$). On the other hand, the power balance gives
\begin{equation}\label{dTEForce}
	\dot{E} = TV - DV,
\end{equation}
where $T$ and $D$ denote thrust and drag, respectively. Combining \eqref{dTEDefine} and \eqref{dTEForce}, we define the TECS-referenced (energy) acceleration
\begin{equation}\label{aTEDefine}
	a_{TE} := \dot{V} - g\frac{V_z^i}{V} = \frac{T}{m} - \frac{D}{m}.
\end{equation}

This quantity captures the net acceleration produced by non-gravitational forces and naturally accounts for the coupling between speed and altitude changes. With negligible wind, $V\approx V_a$, and under the small-$\alpha$, small-$\beta$, and quasi-steady assumptions, the drag coefficient can be treated as approximately constant~\cite{anderson2011ebook}. Neglecting induced-drag variations, the drag is approximated by
\begin{equation}\label{drag}
	D \approx \frac{1}{2}\rho S C_{D} V_a^2.
\end{equation}

Therefore, for a fixed thrust command, substituting \eqref{drag} into \eqref{aTEDefine} yields an approximately linear relationship between $a_{TE}$ and $V_a^2$, which can be expressed as
\begin{equation}\label{fit}
	a_{TE}
	=
	\underbrace{\left(-\frac{\rho S C_D}{2m}\right)}_{k^{V}} V_a^2
	+
	\underbrace{\left(\frac{T_c}{m}\right)}_{b^{V}}
\end{equation}

To identify the mapping in flight, we select several thrust command levels $T_{c,i}$ and collect the corresponding $(V_a^2,a_{TE})$ data. For each thrust level, linear regression yields an estimate of parameters $\hat{k}_i^V$ and $\hat{b}_i^V$ in \eqref{fit}. 

Given the current airspeed $V_a$, these fitted models provide a set of sample pairs relating $a_{TE}$ and $T_c$. We then fit an inverse (command) model of the form
\begin{equation}\label{fit_T}
	a_{TE} = \hat{k}^T T_c + \hat{b}^T,
\end{equation}
which enables online computation of the thrust command needed to achieve a desired energy acceleration:
\begin{equation}
	T_c = \frac{a_{TE,c} - \hat{b}^T}{\hat{k}^T},
\end{equation}
where the command tangential acceleration $a_{TE,c}$ is given by
\begin{equation}
	a_{TE,c} = a_c^t - g\frac{V_z^i}{V},
\end{equation}
with $a_c^t := \mathbf{a}_c^t\cdot \mathbf{e}_x^v$.

The relationship between $T_c and a_{TE,c}$ can be obtained experimentally and may exist bias under varying aerodynamic conditions. 
To compensate for such modeling errors, a simple integral feedback term can be incorporated into the tangential acceleration loop, effectively reducing steady-state discrepancies while preserving the acceleration-level control framework.

Compared with directly mapping $a_c^t$ to thrust, this energy-based method (i) consistently accounts for climbs/descents through the gravity coupling term and (ii) avoids thrust-bench calibration or detailed propulsion modeling, since the mapping is identified directly from flight data.

\subsection{Dynamic Coupling and Priority Handling of Normal and Tangential Accelerations}

Under near-level flight, normal and tangential dynamics are almost decoupled. During climbs or descents, however, gravity strongly influences the tangential channel, and the desired $\mathbf{a}_c^t$ may be infeasible to realize using thrust and drag alone. For example, during a steep, high-speed descent, even saturating the thrust command at $T_c=0$ may be insufficient to produce the requested negative tangential acceleration. In such cases, coupling between normal and tangential channels is unavoidable, and it is necessary to prioritize one channel while respecting actuator limits. We use the standard saturation function
\begin{equation}
	\text{sat}(x, x_{\min}, x_{\max}) =
	\begin{cases}
		x_{\min}, & x < x_{\min}, \\
		x, & x_{\min} \le x \le x_{\max}, \\
		x_{\max}, & x > x_{\max}.
	\end{cases}
\end{equation}

\subsubsection{Normal-Acceleration-Priority Strategy}

In the normal-acceleration-priority mode, the thrust command is saturated as
\begin{equation}
	T_c \leftarrow \text{sat}(T_c, 0, 1),
\end{equation}
while the attitude-rate commands $(p_c,q_c)$ computed from the normal-acceleration mapping remain unchanged. In this mode, the aircraft prioritizes achieving the desired normal acceleration by redirecting the lift vector, and the tangential response is determined by the available thrust. This strategy is suitable for guidance and intercept tasks, where normal maneuverability is more critical than strict airspeed regulation. In extreme cases, overly aggressive normal commands may approach stall boundaries; handling such envelope constraints can be incorporated by additional supervisory logic and is left for future work.

\subsubsection{Tangential-Acceleration-Priority Strategy}

In the tangential-acceleration-priority mode, the identified $T_c$--$a_{TE}$ mapping provides energy-acceleration bounds $a_{TE,\min}$ and $a_{TE,\max}$ corresponding to $T_c=0$ and $T_c=1$ at the current airspeed $V_a$. From the definition of $a_{TE}$ in Eq. \eqref{fit},
\begin{equation}
	\frac{V_z^i}{V} = \frac{a_c^t - a_{TE}}{g}.
\end{equation}

Under the small-$\alpha$ assumption, $\sin\theta \approx \mathbf{V}\cdot \mathbf{e}_z^i / V$, leading to a pitch-angle interval that satisfies the tangential-acceleration constraint:
\begin{equation}
	\begin{aligned}
		\theta_{\min} &= \arcsin \frac{a_c^t - a_{TE,\max}}{g}, \\
		\theta_{\max} &= \arcsin \frac{a_c^t - a_{TE,\min}}{g}.
	\end{aligned}
\end{equation}

We generate pitch-rate bounds via proportional control,
\begin{equation}
	\dot{\theta}_{c,\min} = k_\theta(\theta_{\min} - \theta), \quad
	\dot{\theta}_{c,\max} = k_\theta(\theta_{\max} - \theta),
\end{equation}
where $k_\theta$ is the tuning gain, and by neglecting roll/yaw coupling in the pitch channel, set
\begin{equation}
	q_{c,\min} = \dot{\theta}_{c,\min}, \quad
	q_{c,\max} = \dot{\theta}_{c,\max}.
\end{equation}

Finally, we saturate the pitch-rate and thrust commands:
\begin{equation}
	q_c \leftarrow \text{sat}(q_c, q_{c,\min}, q_{c,\max}), \quad
	T_c \leftarrow \text{sat}(T_c, 0, 1).
\end{equation}

This strategy prioritizes tangential behavior by adjusting the flight-path angle when necessary. For instance, during a prolonged descent where the requested deceleration cannot be achieved by thrust reduction alone, the controller pitches up to reduce airspeed, which is consistent with the gliding deceleration behavior of fixed-wing aircraft. Depending on mission requirements, a supervisory layer can switch between the normal-acceleration-priority and tangential-acceleration-priority strategies (e.g., prioritize normal acceleration during aggressive guidance maneuvers and prioritize tangential acceleration during energy management).

\section{Experimental Validation and Results}

\begin{figure}[tb!]
	\centering
	\includegraphics[width=0.4\textwidth]{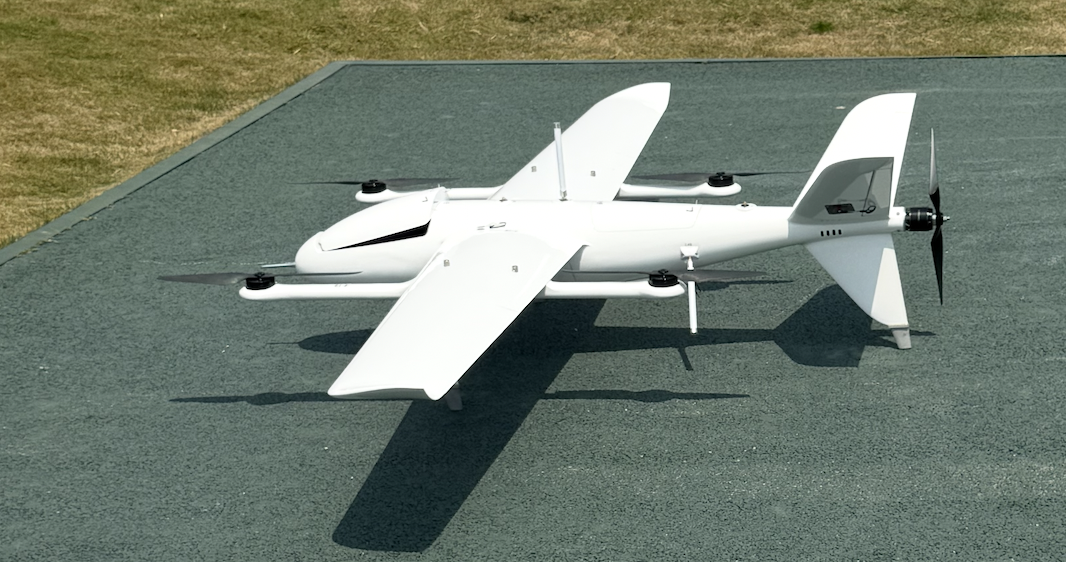}
	\caption{SkyFury VTOL fixed-wing Platform.}
	\label{vtol}
\end{figure}

To evaluate the engineering feasibility and control performance of the acceleration-level method proposed in Section~\ref{sec:accel_control}, a series of outdoor flight experiments were conducted on SkyFury (see Fig. \ref{vtol}), a vertical takeoff and landing (VTOL) fixed-wing UAV platform. The experiments include: (i) in-flight identification of the mapping between the normalized thrust command and the energy acceleration, (ii) acceleration-tracking response tests under commanded normal and tangential accelerations, and (iii) a proportional-navigation (PN) guidance demonstration on a virtual target-point intercept task.

\subsection{System Architecture and Experimental Setup}

The flight experiments were conducted using a VTOL fixed-wing UAV as the experimental platform. Key physical parameters of the UAV are listed in Table~\ref{vtol_params}. The UAV is equipped with a PX4 autopilot as the low-level flight control system, which provides stable tracking of attitude-rate commands and the normalized thrust command $T_c$.

\begin{table}[tbp]
	\centering
	\caption{Key parameters of the VTOL SkyFury}
	\begin{tabular}{l c}
		\hline
		Parameter & Value \\
		\hline
		Wingspan & 2.50 m \\
		Length & 1.26 m \\
		Empty weight & 6.58 kg \\
		Battery weight & 4.72 kg \\
		Maximum payload & 6.92 kg \\
		Max. takeoff weight (MTOW) & 13.50 kg \\
		Cruise speed & 19--30 m/s \\
		Maximum speed & 41.7 m/s \\
		Endurance (at 12.5 kg takeoff weight) & 3 h 20 min \\
		\hline
	\end{tabular}
	\label{vtol_params}
\end{table}

The missions were executed using a hybrid \texttt{MISSION--OFFBOARD--MISSION} mode of the PX4 flight control system. During takeoff and waypoint navigation, the UAV operated autonomously in \texttt{MISSION} mode under PX4 control. Upon reaching the predefined experimental area, the flight mode was switched to \texttt{OFFBOARD}, where the onboard high-level controller sent attitude-rate commands $(p_c,q_c)$ and a normalized thrust command $T_c \in [0,1]$ in real time to validate the proposed acceleration-level control algorithms. After completing the experiments, the UAV returned to \texttt{MISSION} mode to autonomously navigate back and land.

The UAV's navigation system is equipped with an onboard sensor suite that includes an inertial measurement unit (IMU) for attitude angles, angular rates, and body accelerations; a pitot-tube airspeed sensor for $V_a$; and an RTK-capable GNSS receiver for position and velocity. State variables and control commands were exchanged between the onboard computer and the flight control system via ROS2, enabling integrated control and synchronized data logging.

\subsection{Flight 1: Thrust--Energy Acceleration Mapping}

\begin{figure}[tpb]
	\centering
	\includegraphics{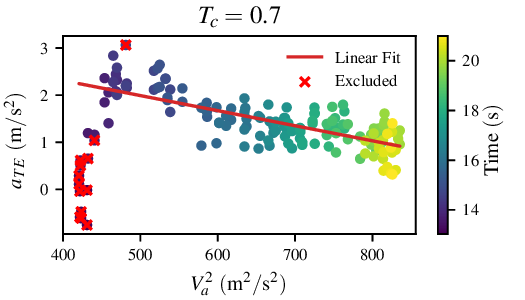}
	\caption{Relationship between $V_a^2$ and $a_{TE}$ at $T_c = 0.7$.}
	\label{ta}
\end{figure}
\begin{figure}[tpb]
	\centering
	\includegraphics{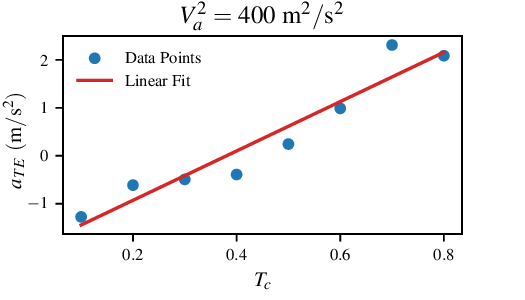}
	\caption{Relationship between $a_{TE}$ and $T_c$ at $V_a = 20\,\mathrm{m/s}$.}
	\label{Tafit}
\end{figure}

To establish the mapping between the thrust command $T_c$ and the energy acceleration $a_{TE}$ in flight, calibration flights were first conducted. During the experiments, the thrust command was set within
\begin{equation}
	T_c \in [0.1,\,0.8],
\end{equation}
with an increment of $0.1$. A total of eight discrete thrust levels were applied, and each level was maintained for $2~\mathrm{s}$ to reduce the transient effect of propeller speed changes during thrust switching. To ensure flight safety and sufficient airspeed margin, thrust levels were applied in an alternating sequence rather than monotonically increasing.

During the calibration, a simple attitude stabilization law was used:
\begin{equation}
	p_c = -k_\phi \phi, \qquad
	q_c = -k_\theta (\theta - \theta_0),
\end{equation}
which kept the aircraft near level flight and reduced the influence of attitude variations on the identification results.

Under fixed thrust conditions, the energy acceleration response was recorded at different airspeeds. For each thrust level, linear regression of $a_{TE}$ versus $V_a^2$ yields the parameters $\hat{k}_i^V$ and $\hat{b}_i^V$ in \eqref{fit}. Fig.~\ref{ta} shows an example at $T_c=0.7$; after removing transient samples, the fitted line yields $\hat{k}_7^V = -0.0032$ and $\hat{b}_7^V = 3.5912$. Using the set of fitted models, the $T_c$--$a_{\mathrm{TE}}$ relationship can be reconstructed at an arbitrary airspeed, and a linear fit of $a_{TE}$ versus $T_c$ is obtained as in Fig.~\ref{Tafit}. This mapping is then used online to compute $T_c$ from a desired $a_{TE,c}$.

\subsection{Flight 2: Normal--Tangential Acceleration Tracking}

\begin{figure}[!tb]
	\centering
	\includegraphics{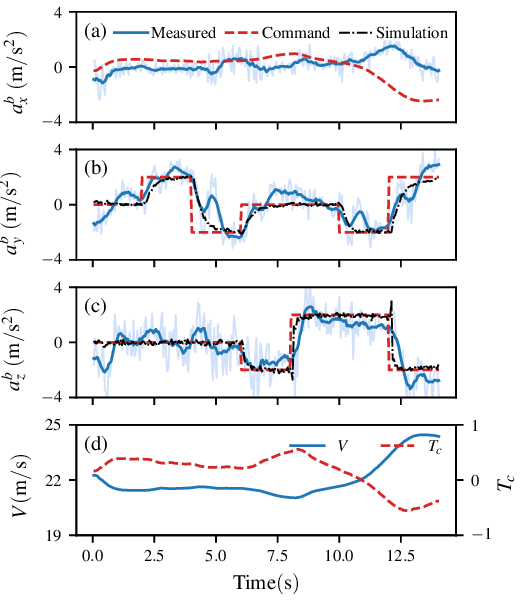}
    \caption{Results of the coupled normal--tangential acceleration response experiment.
	(a)--(c) Command and measured body-frame accelerations along the $x$, $y$, and $z$ axes. 
	The light-blue curves denote raw measured signals, while the dark-blue curves represent moving-average filtered data over a 10-sample window.
    The black dash-dotted curves show the corresponding responses obtained from Gazebo simulation under the same command sequence.
	(d) Ground speed and normalized thrust command $T_c$.}
	\label{na}
\end{figure}

After identifying the thrust--energy acceleration mapping, acceleration-tracking flights were conducted to validate the proposed method under realistic conditions. Piecewise step commands were applied in the normal channel to assess transient response and steady-state tracking. 
In parallel, a simple speed regulation loop generated the tangential acceleration command,
\begin{equation}
	a_c^t = k_V (V_c - V),
\end{equation}
where $V_c$ is the target ground speed and $k_V>0$ is an empirically chosen gain (units: $\mathrm{s^{-1}}$). When the commanded tangential behavior became infeasible due to actuator limits, the controller followed the normal-acceleration-priority strategy described in Section~\ref{sec:accel_control}.
To provide a baseline reference without wind disturbances and sensor noise, the same command sequence was also evaluated in the Gazebo simulation environment. The simulation results are plotted together with the flight-test data for comparison.

Fig.~\ref{na} shows the commanded and measured accelerations during a representative run. After a brief transient, the UAV achieves stable tracking of the commanded normal acceleration steps, indicating that the proposed method can effectively realize acceleration commands using only body-rate and thrust interfaces. The smoother responses with simulation indicates that the fluctuations observed in flight are mainly caused by wind disturbances and sensor noise.
During $0$--$11~\mathrm{s}$, the speed regulation loop maintains a nearly constant ground speed by adjusting the thrust command. After $11~\mathrm{s}$, the thrust command reaches saturation and the vehicle prioritizes the normal acceleration; correspondingly, the ground speed deviates from the reference, which is consistent with the designed priority mechanism.

\subsection{Flight 3: PN Guidance Application}
\begin{figure}[!tb]
	\centering
	\includegraphics{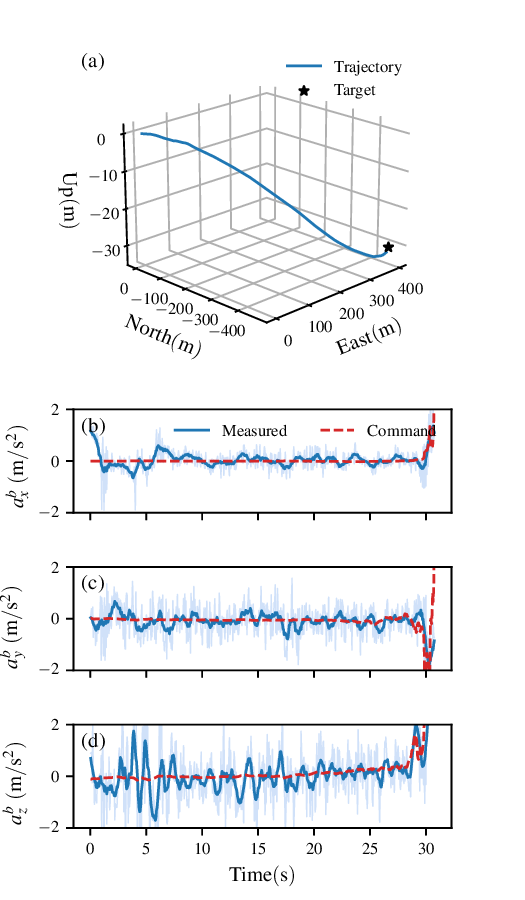}
	\caption{Results of the proportional navigation flight experiment. 
		(a) Flight trajectory and virtual target point. Miss distance: 0.58 m.
		(b)--(d) Command and measured body-frame accelerations along the $x$, $y$, and $z$ axes. 
		Light-blue curves represent raw measured signals, and dark-blue curves show moving-average filtered data over a 10-sample window.}
	\label{PN}
\end{figure}

To demonstrate that acceleration-commanded guidance laws can be applied in real guidance applications, a PN guidance experiment was conducted. PN commands a normal acceleration proportional to the line-of-sight (LOS) rate:
\begin{equation}
	\mathbf{a}_c^n = N V_{cl} \dot{\lambda},
\end{equation}
where $N$ is the navigation constant, $V_{cl}$ is the closing speed to the target, and $\dot{\lambda}$ is the LOS angular rate.

In the experiment, the target was specified as a virtual waypoint located 600~m horizontally and 30~m vertically from the guidance start point. The UAV was commanded to regulate its ground speed to $20~\mathrm{m/s}$ using the same tangential command generation as in the previous experiment. During flight, the onboard computer continuously computed the relative target position to obtain $\lambda$ and $\dot{\lambda}$, generated the normal acceleration command via PN, and then converted it into attitude-rate commands and a normalized thrust command using the proposed acceleration-level outer loop.

The results are shown in Fig.~\ref{PN}. The UAV successfully intercepted the virtual target with a miss distance of 0.58~m, and the measured accelerations closely follow the commanded profiles. These results support the practical feasibility of the proposed acceleration-level outer loop for real-world implementation of acceleration-based guidance laws on fixed-wing UAVs.

\section{Conclusions}
This paper proposed a practical acceleration-command realization layer for fixed-wing UAVs, enabling acceleration-commanded guidance laws to be deployed on standard autopilots through body-rate and normalized-thrust interfaces. By decomposing the commanded acceleration into normal and tangential components, the proposed outer loop maps the normal command to roll- and pitch-rate inputs that shape the lift vector, and maps the tangential command to a thrust input via a TECS-inspired energy formulation whose parameters are identified directly from flight data. Flight experiments on a VTOL fixed-wing platform validated accurate acceleration tracking and demonstrated PN-based target-point intercept using only body-rate and normalized thrust interfaces.

Future work will relax the small-angle assumptions and explicitly account for wind and broader flight-envelope constraints (e.g., stall margin and actuator rate limits). In addition, we will investigate formal stability/robustness guarantees and supervisory switching between normal- and tangential-priority modes to better accommodate diverse mission objectives.

\bibliographystyle{IEEEtran} 

\bibliography{ref}

\end{document}